# Decision method choice in a human posture recognition context

**Stéphane Perrin, Eric Benoit, Didier Coquin**

LISTIC Lab - Université Savoie Mont Blanc, Annecy, France
{stephane.perrin, eric.benoit, didier.coquin}@univ-smb.fr

**Abstract.** Human posture recognition provides a dynamic field that has produced many methods. Using fuzzy subsets based data fusion methods to aggregate the results given by different types of recognition processes is a convenient way to improve recognition methods. Nevertheless, choosing a defuzzification method to implement the decision is a crucial point of this approach. The goal of this paper is to present an approach where the choice of the defuzzification method is driven by the constraints of the final data user, which are expressed as limitations on indicators like confidence or accuracy. A practical experimentation illustrating this approach is presented: from a depth camera sensor, human posture is interpreted and the defuzzification method is selected in accordance with the constraints of the final information consumer. The paper illustrates the interest of the approach in a context of postures based human robot communication.

## 1 Introduction

Human posture recognition provides a dynamic field that has produced many methods [Mitra and Acharya 2007] that commonly perform a measurement process in order to convert the physical state of an entity into an information entity. Although numerical values are commonly used to represent measurement results, it is now admitted that some applications manipulate symbolic values or linguistic terms better.

In this paper, we concentrate first on recognizing human posture from upper limb posture detection. The two parts (arm and forearm) are expressed using linguistic terms. The corresponding symbolic values result from numeric angle measurements of the human body provided by a depth camera sensor, in our case an Asus Xtion Pro Live. We also use linguistic terms to express postures from symbolic values of the upper limb parts.

Then we focus on the decision process performed from the symbolic values, which are commonly expressed in terms of mass. Fusion methods are particularly effective here and are used to manipulate mass [Dubois and Prad 1988]. They are generally used in the decision system as they can formalize decision criteria from fuzzy information, especially when evidence theory is used. Finally, we propose to express the needs of the information consumer, which are viewed as constraints on the decision process. This allows to choose and compute the decision, i.e. action to be done, in accordance with the consumer's needs and use.

Posture recognition carried out by a human is complex and the decision process takes into consideration criteria like hesitation between known postures or belief in a recognition posture. Fuzzy logic facilitates the manipulation of these concepts. In the case of automatic and artificial processes for human posture recognition, we show that results of fusion methods associated to the evidence theory are a good way to express consumer criteria, which are finally used to select the defuzzification method.

This approach, illustrated in section 4, is used to give orders to a humanoid NAO robot developed by the Aldebaran Company, from human posture recognition. The system learns a set of reference postures, each one corresponding to an order chosen to be transmitted to the robot. For example: the user wants the robot to execute an order only if the recognition of the corresponding human posture is sure, i.e. with low uncertainty or not ambiguous with other possible reference postures. Another example: a posture corresponds to an emergency action. In this case, if this posture is recognized with enough certainty, the corresponding order is chosen even if another posture is recognized at the same time, but with a greater certainty.

In section 2, we present posture modeling using fuzzy representation. Section 3 details the posture recognition process.

In section 4 the decision process is described using distance criteria from learning reference postures. This approach allows for defining and formalizing user constraints like recognition quality or threshold distance of posture detection. Facilitating the decision process based on mass transfers inside fuzzy representation of human posture is introduced.

User constraints are thrown back to the decision system which is now able to choose the robot order to be applied, in accordance to user expectation.

## 2 Fuzzy logic for human posture modeling

This section goes back over the concepts of fuzzy nominal scale and metrical scale presented in [Benoit and Foulloy 2013] that are used to represent postures.

### 2.1 Fuzzy representation: fuzzy nominal scale

This section goes back over the concepts of fuzzy nominal scale and metrical scale presented in [Benoit and Foulloy 2013] that are used to represent postures. A fuzzy nominal scale is defined as the link between a quantity manifestation and its representation by a fuzzy subset of linguistic terms also called Lexical Fuzzy Subset and denoted LFS in this paper. It gives a formal framework to the fuzzy mapping process presented in [Mauris et al. 1994; Benoit and Foulloy 2002]. Its has been already shown that the fuzzy nominal scales map an equality relation on lexical term to a fuzzy similarity on LFSs. In [De Baets and Mesiar 2002], De Baets and Mesiar presented multiple possible fuzzy relations that can be used to extend the concept of fuzzy nominal scales. We especially point out the introduction of a new distance between LFSs in [Allevard et al. 2007]. This distance, called transportation distance and inspired by the Earth Mover Distance in [Rubner et al. 2000], is linked with an arbitrary predefined distance between lexical terms and extends it by the way of a metrical scale as presented in [Benoit and Foulloy 2013].

Let $d_S$ be an arbitrary defined distance between lexical terms. This distance reflect a knowledge related to the semantic of the lexical terms. It can be experimentally defined using a calibration process. Another way is to use this distance to include an knowledge on the relation between lexical terms. For example, the distance between two terms representing similar entities is arbitrarily defined as small. The transportation distance $d'_S$ between LFSs is then deduced from $d_S$ such that the coincidence on singletons is respected.

Working with a metrical scale gives much more possibilities than working with a nominal scale even if this one is a fuzzy nominal scale. Indeed the fuzzy nominal scale preserves only the similarity during the measurement: if 2 entities are similar then their representations (as LFSs) throw a fuzzy nominal scale are similar. A metrical scale preserves also the comparisons of distances between entities. This means that the distance comparison can be used on the set of LFSs.

In this paper, we propose to transpose the concept of tolerance interval to the set of LFSs. The chosen solution to implement this is to create a tolerance volumes defined with the transportation distance.

Fuzzy scale provides a tool to represent a posture with an LFS, but gives no semantic to the membership degree of each term to this fuzzy subset. In this study, we decided to choose a weak semantic by the interpretation of the fuzzy membership degrees as belief masses as defined in the TBM (Transferable Belief Model) [Smets 2000]. According to this semantic, each LFS is termed as Basic Belief Assignment i.e. the membership degree of a LFS is interpreted as a unitary mass distributed on the singletons.

**2.2 Human posture representation**

There has been a lot of focus on the development of natural interfaces using human communication modalities in the human system interaction field. One of these is communication through postures that can be described with words. Fig. 1 (left) illustrates the posture "hello" in a 2D representation of the body detected by the depth sensor and the measured joints. This particularity of postures makes it possible to describe them with fuzzy nominal scales.

The depth camera sensor (the Asus Xtion Pro Live) produces posture detection using NiTE library. The result of this process is a set of 3D joint positions that we call the measured skeleton.

We illustrate our approach with a right upper limb detection posture. To express the human posture representation, we decompose arm and forearm representation into 2 corresponding intermediate high-level representations which are expressed using words. Human posture and high-level representations of arm and forearm measurement are linked by rules as shown below.

Each body part (i.e. the arm and forearm) are expressed using words. The angles pertaining to the arm and forearm are computed from the joint positions of the detected skeleton. Lexical sets are chosen to have simple and easily understandable descriptions. Each part (i.e. the arm and forearm) is characterized by the 2 measured angles from the position of the joints, as illustrated in fig. 1. We also use lexical fuzzy subsets as shown below.

In this section, we present arm forearm modeling from respective angles measurement of skeleton. LFS are used. Arm and forearm posture definition rules are detailed. Then, right limb modeling is detailed using LFS too and corresponding rules are detailed too.

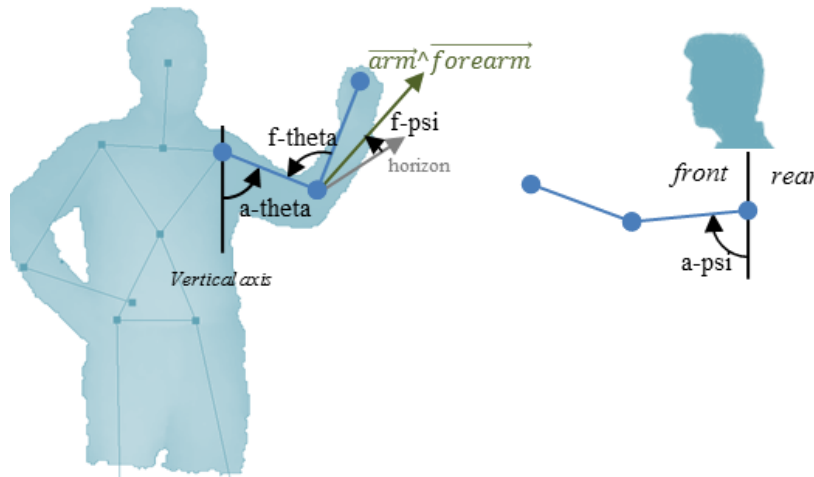

**Fig. 1** 2D representation of 3D measurements of joints for the "hello" posture (left) and used angles from skeleton (right).

1) *Arm modeling*

For arm angle measurement, we use *a-tetha* angle and *a-psi* angle. The *a-tetha* angle corresponds to the angle formed from the vertical to the arm, in the plane of the human body. *a-psi* is the angle formed from the vertical to the arm when the arm is level in front of the human's body. The LFS of *a-theta* is *{down, horizon, up}* and the LFS of *a-psi* angle is *{rear, outside, front, inside}*. Fig. 2 illustrates the *a-theta* and *a-psi* angle lexical posture determination from respective angle measurements. In this step, we present the concept of modal angle values that define representative angles. A modal angle is defined such that the LFS obtained by the fuzzy linguistic description is a singleton.

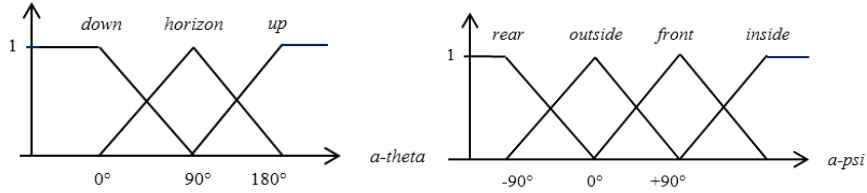

**Fig. 2** Sub-lexical for arm posture from the *a-theta* angle value and the *a-psi* angle value.

Then, the arm posture is represented by the combination of these 2 lexical subsets, corresponding to the 2 *a-tehta* and *a-psi* angles.

The arm corresponding variable takes its values in the lexical set $L_{arm}$ (1):

$$L_{arm} = \{down, front, up, outside, rear, inside\} \qquad (1)$$

Table 1 (left) gives the arm position rules from these lexical subsets.

**Table 1.** Arm and forearm posture definition rules

| arm | | a-tetha | | |
|---|---|---|---|---|
| | | **down** | **horizon** | **up** |
| a-psi | **rear** | down | rear | up |
| | **outside** | down | outside | up |
| | **front** | down | front | up |
| | **inside** | down | inside | up |

| forearm | | f-tetha | | |
|---|---|---|---|---|
| | | **close** | **middle** | **open** |
| f-psi | **vertical** | close | vmiddle | open |
| | **horizontal** | close | hmiddle | open |

*2) Forearm modeling*

A similar process of arm modeling is done for the forearm part. 2 angles are used: *f-tetha* which corresponds to the angle formed by the forearm axis and arm axis and *f-psi,* which corresponds to the position relative to the horizon. The fuzzification process is the same as the arm part and is not detailed in this paper. The forearm corresponding variable takes its values in the lexical set $L_{forearm}$ (2):

$$L_{forearm} = \{open, vclose, hclose, vmiddle, hmiddle\} \qquad (2)$$

Table 1 (right) represents the position rules from the lexical subset corresponding to *f-theta* and *f-psi*.

### 3) Right upper limb modeling

The posture is represented by the combination of an arm and a forearm posture and is also defined by words. Each word represents a human modal posture. Each modal posture of the right upper limb corresponds to a set of modal postures of each part of the limb.

The lexical set of postures is defined by a list of modal postures $L_{P\text{-}mod}$ (table 2).

Using rules, each modal posture made by upper limb postures is defined in the system. This learning step is based on the sub-posture entries: the arm and forearm values. Table 2 illustrates rules defining modal postures from lexical subsets defined in (1) and (2).

It is noted that the example in table 2 illustrates a 2D dimension case due to the *2* linguistic variables (corresponding to the arm and forearm). It is possible to generate the principle to *n* dimensions.

The linguistic set of modal postures is given in (3):

$$L_{P\text{-}modal} = \{front, outside, inside, down, up, rear, frontfolded, outsidefolded, insidefolded, downfolded, upfolded, rearfolded, fronthmiddle, outsidehmiddle, insidehmiddle, downhmiddle, uphmiddle, rearhmiddle, frontvmiddle, outsidevmiddle, insidevmiddle, downvmiddle, upvmiddle, rearvmiddle\} \quad (3)$$

**Table 2.** Modal posture definition rules

|  |  | arm |  |  |  |  |  |
|---|---|---|---|---|---|---|---|
|  |  | *down* | *front* | *up* | *outside* | *rear* | *inside* |
| forearm | *open* | down | front | up | outside | rear | inside |
|  | *close* | downfolded | frontfolded | upfolded | outsidefolded | rearfolded | insidefolded |
|  | *hmiddle* | downhmiddle | fronthmiddle | uphmiddle | outsidehmiddle | rearhmiddle | insidehmiddle |
|  | *vmiddle* | downvmiddle | frontvmiddle | upvmiddle | outsidevmiddle | rearvmiddle | insidevmiddle |

The set of all modal postures $L_{P\text{-}mod}$ can be directly the set of reference postures i.e. known postures to be detected. This particular situation is presented in [Perrin et al. 2015]. In this paper, reference postures are expressed from modal postures. This additional step allows for the application designer to distinguish modal postures from reference postures. Therefore, modal posture definition rules (table 2) are fixed independently from reference postures. In the next sub-section, reference posture modeling is presented based on modal posture representation.

### 2.3 Human reference posture representation

Reference postures are the postures to be detected and are known by the system during the learning process. This section presents their representation from modal postures. Each modal posture is known by a word in the system. Reference postures can be expressed in terms of mass on known modal postures. The mass criterion is computed from the reference posture and each known modal posture. Hence, a reference posture $P_{ref}$ is written as:

$$P_{ref} = (L_{P\text{-}mod,1}, L_{P\text{-}mod,2}, L_{P\text{-}mod,3}, \ldots, L_{P\text{-}mod,n}) \qquad (4)$$

where $n$ is the (maximum) number of modal postures (contained in $L_{P\text{-}mod}$). Note that terms associated with no-neighbor modal postures, if they appear, are set to 0.

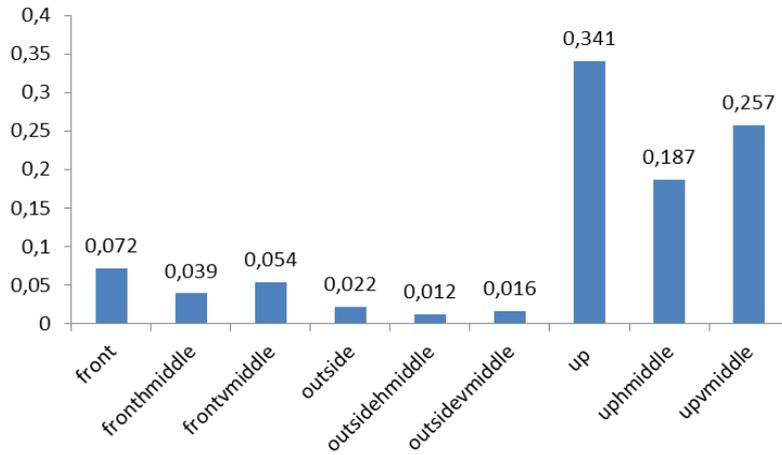

**Fig. 3** *"stop" reference posture representation*.

Table 3 illustrates the "stop" reference posture. In this example, the reference posture is given in table 3 and fig. 3.

**Table 3.** "stop" reference posture determination

|   |   | arm | | | | | |
|---|---|---|---|---|---|---|---|
|   |   | **down** | **front** 0.1653 | **up** 0.7847 | **outside** 0.0501 | **rear** | **inside** |
| **forearm** | **open** 0.4348 | down | front 0.07187 | up 0.3411 | outside 0.02178 | rear | inside |
|   | **close** | downfolded | frontfolded | upfolded | outsidefolded | rearfolded | insidefolded |
|   | **hmiddle** 0.2378 | downhmiddle | fronthmiddle 0.03930 | uphmiddle 0.1866 | outsidehmiddle 0.01191 | rearhmiddle | insidehmiddle |
|   | **vmiddle** 0.3274 | downvmiddle | frontvmiddle 0.05412 | upvmiddle 0.2569 | outsidevmiddle 0.01640 | rearvmiddle | insidevmiddle |

The next section presents a detection posture process.

## 3 Decision process under user constraints

In our application context, the decision process has to choose an action from a human posture recognition result, which is expressed using fuzzy representation. This allows us to express the detection finely: several postures can be detected from one human posture. At the end of the configuration process, posture knowledge is made up of a lexical set, and a set of reference postures. Each reference posture is defined by a fuzzy subset interpreted as a mass distribution on the lexical set [Benoit et al. 2015; Allevard et al. 2005]. A measurement result is also defined as represented by a fuzzy subset interpreted as another mass distribution on the same lexical set. The decision process then acts as a defuzzification.

This fluency of information carried by fuzzy representation is conventionally employed cases using multi-sensor detection. In this paper, we focus on its interests in order to automatize the choice of action to be done from one human posture recognition. The imprecision carries information that represents possible confusion, hesitation or belief of detected postures. The output of the decision process stems from the defuzzification process.

Several reference postures are detected from one human posture detection. In this experiment context, we consider that each reference posture corresponds to an action to be executed by the NAO robot. For each reference posture $L_{P\text{-}ref,i}$ is a corresponding action named $A_i$.

In our first basic approach, the decision is given by the closest reference posture obtained by the computation of the transportation distance on the LFS space, as presented in section II.A., between the measured posture $P_{detect}$ and the reference postures. This is the case in case 1 (fig. 4) where the decision result is a non-empty set of reference postures.

Our second approach considers the uncertainty of the measurement process predominantly coming from camera process detection. The uncertainty estimation is complex in this case and is not presented in this paper. In all cases, epistemic uncertainty, i.e. relative to posture definition, is more important and taken into account by the consumer. We consider that the consumer accepts a tolerance volume around each reference posture on the LFS space, as illustrated in fig. 4. This tolerance volume plays the same role for the LFS space as the tolerance interval for numerical spaces. It is defined by the distance between a measured posture and the reference posture, which is recognized for having the measured posture inside its tolerance volume. In case 2 presented in fig. 4, the measured posture is outside the tolerance volumes of all reference postures and the decision result is an empty set. In case 3 the tolerance volume around $P_{ref,3}$ is large enough to include the measured posture. In this approach the tolerance volumes do not overlap, so the decision result is an empty set or a singleton.

In our third approach, the consumer wants to take into account the possible confusion between several possible postures, which is given by the overlapping of their respective tolerance volume that depends on the distance between the refer-ence postures and on the distance that characterizes the tolerance volume. Case 4 illustrates this situation: set {P$_{ref,1}$, P$_{ref,2}$} is chosen as the partial decision.

Another constraint is added in order to perform the decision. The nature of the robot's action associated to the command posture is able to influence the decision process. To illustrate this, we consider two classes of actions (each corresponding to a reference posture): one class contains "classical" actions and the second "emergency" actions, corresponding respectively to a "classical" reference posture and an "emergency" reference posture. So when an "emergency" reference posture is detected even with a lower mass than a classical detected reference posture, the selected decision is the corresponding "emergency" action. From this consideration, either the nearest emergency reference posture is selected – see case 5 of fig. 4; or the consumer constraint reduces the set of possible recognized postures to the *emergency* related one. This situation is similar to case 2 approach, but with corresponding emergency postures only.

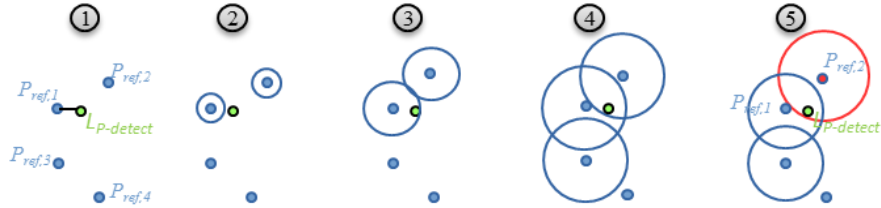

**Fig. 4** Illustration of tolerance volume and the impact on decision.

Let's look at an applicative example on posture recognition: 3 reference postures named *pointing*, *standing* and *protect* are performed and measured, then translated into LFSs on $L_{P\text{-}modal}$. $P_{pointing}$, $P_{standing}$ and $P_{protect}$ to denote the 3 reference postures. These are illustrated in fig 7. A $d_S$ distance is defined on the lexical set of modal postures $L_{P\text{-}modal}$. This distance is arbitrarily chosen in order to respect some constraints:

- The larger distance between 2 modal terms is arbitrarily fixed to 3.0.
- The smaller distance between 2 modal terms related to the same elbow angle, therefore only related to the different shoulder angles, is fixed to 1.0.
- The smaller distance between 2 modal terms related to the same shoulder angle, therefore only related to the different elbow angles, is fixed to 0.5.

Then $d_S$ is extended to a $d'_S$ distance on the LFSs on $L_{P\text{-}modal}$, as presented in 2.1. First the distance between reference postures is computed:

$d'_S(P_{pointing}, P_{standing}) = 0.129$
$d'_S(P_{pointing}, P_{protect}) = 1.8695$
$d'_S(P_{standing}, P_{protect}) = 1.9225$

As expected the *pointing* and *standing* postures are closer to each other than to the *protect* posture. A new $P_1$ posture to be recognized is acquired then fuzzified to produce an LFS. In this example, we take a *stop* posture (see fig. 6).

The strategies for the decision are as follows:
- The closest reference posture: this strategy is not recommended when the set of reference postures is small.
- Non overlapping tolerance volumes: we define a tolerance distance for each reference posture:
  $dtolerance_{pointing} = 0.07$
  $dtolerance_{standing} = 0.05$
  $dtolerance_{protect} = 0.5$
- Possible overlapping tolerance volumes, and the *protect* posture is considered as critical.
  $dtolerance_{pointing} = 0.10$
  $dtolerance_{standing} = 0.08$
  $dtolerance_{protect} = 1.0$

The $P_1$ distances to the references are computed.
$d'_S(P_1, P_{pointing}) = 1.8475$, $d'_S(P_1, P_{standing}) = 1.9005$, $d'_S(P_1, P_{protect}) = 0.584$

In this case, the final decision depends on the chosen strategy. The last one recognizes the *protect* posture even if the measured one is not so close. With the second strategy, no posture is recognized.

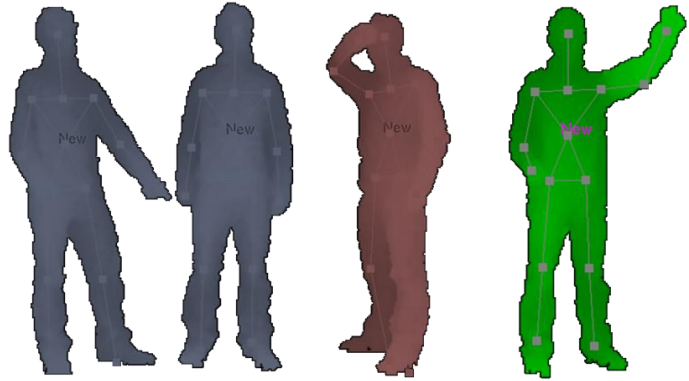

**Fig. 6** 3 reference postures: pointing, standing and protecting and the detected posture (stop).

## 4 Perspectives

In the future our work will consider several posture recognition sensors and types of sensors. This will illustrate the interest of the multi-modal fusion approach which will make it possible to manage conflicting information. In addition, we will look into how to consider human gesture and its repetition from a temporal point of view, which will also allow us to study how fast it is done.

It would be worth studying the impact quality of the recognition process using modal information (i.e. posture) as reference. Indeed, on the one hand, not using additional reference representations simplifies the system but the learning step is dependent on the amount of known information (postures) to be detected. On the other hand, representing reference information (postures) to be detected facilitates adding new reference information. But fixed modal postures have to be chosen in order for the system to be able to represent new references.

# References

[Allervard et al. 2005]
Allevard T, Benoit E, Foulloy L (2005) Dynamic gesture recognition using signal processing based on fuzzy nominal scales. In: Measurement, vol. 38, no. 4, pp. 303-312

[Allevard et al. 2007]
Allevard T, Benoit E, Foulloy L 2007) The transportation distance for fuzzy descriptions of measurement. In: Metrology and Meas. Syst., vol. XIV no. 1, pp. 25-37

[Benoit and Foulloy 2002]
Benoit E, Foulloy L (2002) Fuzzy nominal scale. In: IMEKO TC7 Symposium, Cracow, Poland, pp. 21–25.

[Benoit and Foulloy 2013]
Benoit E, Foulloy L (2013) The role of fuzzy scales in measurement theory. In: Measurement, vol. 46, no. 8, pp. 2921-2926.

[Benoit et al. 2015]
Benoit E, Perrin S, Coquin D (2015) Body posture measurement in a context of example-based teaching. In: Journal of Physics: conference series, vol. 588, 012055, Madeira, Portugal

[De Baets and Mesiar 2002]
De Baets B, Mesiar R (2002) Metrics and T-equalitie. I: Journal of Mathematical Analysis and Applications, 267, pp. 531–547

[Dubois and Prad 1988]
Dubois D, Prade H (1988) Representation and combination of uncertainty with belief functions and possibility measures. In: Computational Intelligence, vol. 4, no. 3, pp. 244-264

[Mauris et al. 1994]
Mauris G, Benoit E, Foulloy L (1994) Fuzzy symbolic sensors - from concept to applications. In: Measurement, vol. 12, pp. 357–384.

[Mitra and Acharya 2007]
Mitra S., Acharya T (2007) Gesture Recognition: A Survey. In: IEEE Trans. Sys., Man, and Cybernetics, Part C: Applications and Reviews, vol. 37, no. 3, May 2007, pp. 311-324

[Perrin et al. 2015]
Perrin S, Benoit E, Coquin D (2015) Fusion method choice driven by user constraints in a human gesture recognition context. In: Proc IEEE 8th International Conference on Human System Interactions HIS 2015, Warsaw, Poland, pp 316-321

[Rubner et al. 2000]
Rubner Y, Tomasi C, Guibas L J (2000) The Earth Mover's Distance as a Metric for Image Retrieval. In: International Journal of Computer Vision, November 2000, Volume 40, Issue 2, pp 99-121

[Smets 2000]
Smets, P (2000) Data fusion in the transferable belief model. In: Proc. IEEE Int. Conference on information fusion, fusion 2000, vol. 1, pp. PS21-PS33.